\documentclass[times,twocolumn,final,num]{elsarticle}

\usepackage{prletters}
\usepackage{framed,multirow}

\usepackage{amsmath,amssymb}
\usepackage{mathtools}
\usepackage{graphicx}
\usepackage{subfig}
\usepackage{afterpage}
\usepackage{multirow}
\usepackage{algorithm}
\usepackage{algpseudocode}
\usepackage{latexsym}
\usepackage{dsfont}
\usepackage{physics} 
\usepackage{MnSymbol} 
\biboptions{sort&compress} 

\usepackage{url}
\usepackage{xcolor}
\definecolor{newcolor}{rgb}{.8,.349,.1}

\newcommand{\binimagesymbol}{\ensuremath{L}}
\newcommand{\binimage}[2]{\ensuremath{\binimagesymbol_{#2}\!\left[#1\right]\!}}
\newcommand{\rangeof}[1]{\ensuremath{\mathfrak{#1}}}
\newcommand{\domainof}[1]{\ensuremath{X_{#1}}}

\newcommand{\card}[1]{\ensuremath{\vert #1 \vert}}
\newcommand{\cardofrangeof}[1]{\card{\rangeof{#1}}}
\newcommand{\nlogn}[1]{\ensuremath{#1\log{#1}}}
\newcommand{\nbinlogn}[1]{\ensuremath{#1\log_2{#1}}}
\newcommand{\searchspace}{S}
\newcommand{\bigo}[1]{\ensuremath{\mathcal{O}(#1)}}
\newcommand{\ha}{\ensuremath{H_A}}
\newcommand{\hb}{\ensuremath{H_B}}
\newcommand{\hab}{\ensuremath{H_{AB}}}

\newcommand{\histentryjoint}[1]{\ensuremath{C_{AB}^{#1}}}
\newcommand{\histentrymarga}[1]{\ensuremath{C_{A}^{#1}}}
\newcommand{\histentrymargb}[1]{\ensuremath{C_{B}^{#1}}}

\newcommand{\shiftvar}{\ensuremath{\chi}}
\newcommand{\posvar}{\ensuremath{x}}

\newcommand{\ceil}[1]{\lceil{#1}\rceil}
\newcommand{\Tstrut}[1][2.6ex]{\rule{0pt}{#1}}

\newcommand{\shiftimage}[1]{\ensuremath{T_{\shiftvar}\!\left[#1\right]}}

\DeclareMathOperator*{\argmax}{arg\,max}

\newcommand{\codeurl}{\url{github.com/MIDA-group/globalign}\,}

\newcommand{\algcorr}[2]{\ensuremath{{#1} \thinstar {#2}}} 
\newcommand{\CC}[2]{(\algcorr{#1}{#2})(\shiftvar)}

\journal{Pattern Recognition Letters}

\begin{document}

\ifpreprint
  \setcounter{page}{1}
\else
  \setcounter{page}{1}
\fi

\begin{frontmatter}

\title{Fast computation of mutual information in the frequency domain with applications to global multimodal image alignment}

\author[1]{Johan \snm{\"{O}fverstedt}\corref{cor1}} 
\cortext[cor1]{Corresponding author: }
\ead{johan.ofverstedt@it.uu.se}
\author[1]{Joakim \snm{Lindblad}}
\author[1]{Nata\v{s}a \snm{Sladoje}}

\address[1]{Department of Information Technology, Uppsala University, L\"{a}gerhyddsv\"{a}gen 2, 752 37 Uppsala, Sweden}

\received{1 May 2013}
\finalform{10 May 2013}
\accepted{13 May 2013}
\availableonline{15 May 2013}
\communicated{S. Sarkar}

\begin{abstract}
Multimodal image alignment is the process of finding spatial correspondences between images formed by different imaging techniques or under different conditions, to facilitate heterogeneous data fusion and correlative analysis. The information-theoretic concept of mutual information (MI) is widely used as a similarity measure to guide multimodal alignment processes, where most works have focused on local maximization of MI that typically works well only for small displacements; this points to a need for global maximization of MI, which has previously been computationally infeasible 
due to the high run-time complexity of existing algorithms. We propose an efficient algorithm for 
computing MI for all discrete displacements (formalized as the cross-mutual information function (CMIF)), which is based on cross-correlation computed in the frequency domain. We show that the algorithm is equivalent to a direct method while asymptotically superior in terms of run-time. Furthermore, we propose a method for multimodal image alignment for transformation models with few degrees of freedom (e.g. rigid) based on the proposed CMIF-algorithm. We evaluate the efficacy of the proposed method on three distinct benchmark datasets, of aerial images, cytological images, and histological images, and we observe 
excellent success-rates (in recovering known rigid transformations), overall outperforming alternative methods, including local optimization of MI as well as several recent deep learning-based approaches. We also evaluate the run-times of a GPU implementation of the proposed algorithm and observe speed-ups from 100 to more than 10,000 times for realistic image sizes compared to a GPU implementation of a direct method. Code is shared as open-source at \codeurl.
\end{abstract}


\end{frontmatter}


\section{Introduction}
\label{sec1}

Multimodal image alignment (also known as registration) 
\citep{zitova2003image}, involves finding correspondences between images formed by different imaging techniques or under different conditions. 
The goal is often to enable data fusion of the heterogeneous information of the sources involved. Processes closely related to registration are patch retrieval and template matching.
Multimodal image alignment can be a very challenging problem, due to great dissimilarities of the involved modalities. One area where fully automated approaches have been lacking is in alignment of micrographs, and in particular correlative microscopy, where image modalities are often highly distinct and images contain small/thin structures that are difficult to match without relying on fiducial markers or time-consuming manual intervention \citep{paul2017ec}.

\begin{figure*}[ht]
\centering
\includegraphics[width=\textwidth]{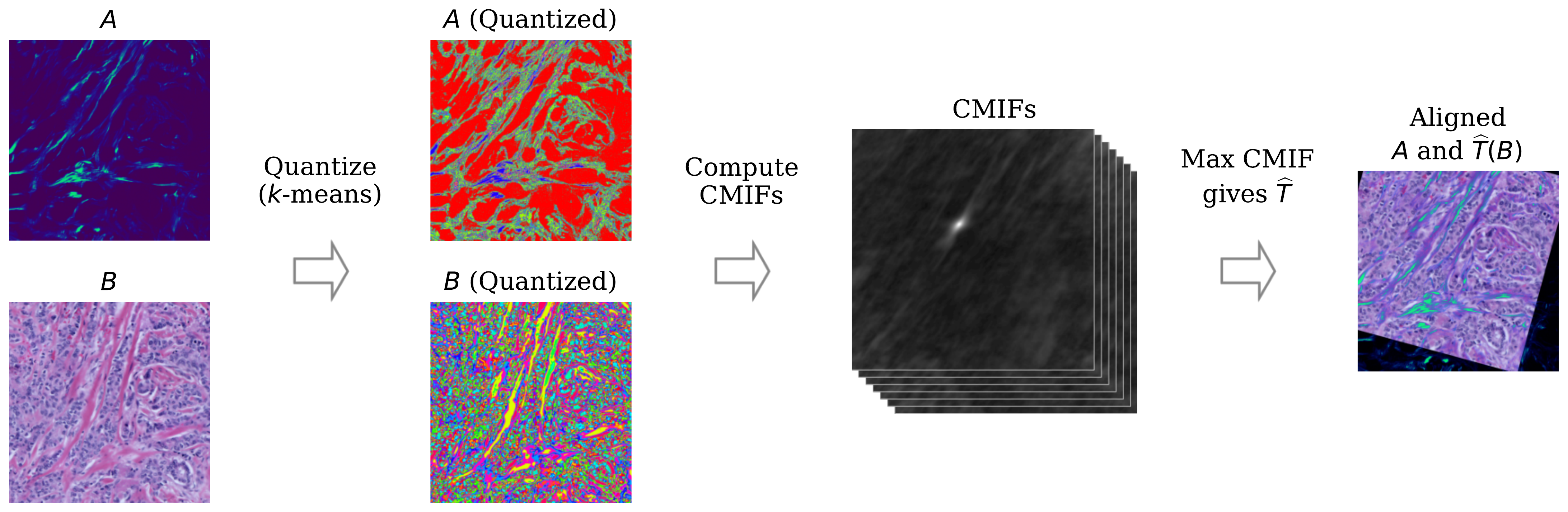}
\caption{Illustration of the main steps of the proposed image alignment method. The input consists of two unaligned histological images from SHG (modality A) and BF imaging (B). They are both quantized into 16 categorical labels using $k$-means clustering (illustrated by pseudo-coloring) followed by computation of dense CMIF maps (MI over all discrete displacements) for each considered rotation angle. The transformation with the max CMIF is identified; it provides the desired alignment. At the end of the pipeline we observe the transformed overlapping images which are successfully aligned.}
\label{fig:mainillustration}
\end{figure*}

Monomodal (or unimodal) alignment can be addressed by a number of existing techniques, often grouped into (i) feature-based methods, such as SIFT \citep{loweObjectRecognitionLocal1999}, where the focus is on finding correspondences between distinct feature points detected in the images, and (ii) intensity-based methods, such as \citep{ofverstedtFastRobustSymmetric2019}, where the alignment is guided by similarity between the whole image functions. Multimodal alignment is more complex and challenging, since similarity in appearance can not be expected for corresponding structures. This reduces the applicability of feature-based methods.  Instead, intensity-based methods guided by local optimization of similarity measures which rely on some statistics, such as mutual information (MI) \citep{viola1997alignment,pluim2003mutualsurvey} or modality independent neighbourhood descriptors \citep{heinrichMINDModalityIndependent2012},  are well-established tools for this task. Another class of methods rely on reducing a multimodal alignment task to a monomodal one, by (typically learning-based) transformation of the image modalities to a common modality \citep{pielawskiCoMIRContrastiveMultimodal2020,lu2021image}, for which high-performance monomodal methods are applicable.

It has been observed that MI exhibits high performance when used as a similarity measure in local optimization frameworks if the displacements to recover are small, but struggles for larger displacements \citep{ofverstedtFastRobustSymmetric2019,lu2021image,pluim2003mutualsurvey}. This highlights the need for fast global optimization of MI. However, existing methods \citep{barrera2010multimodal,shams2010parallel} exhibit run-time complexities which make them unsuited for many practically relevant image sizes. 

We here present two algorithms. The first algorithm, and main contribution of this work, efficiently computes MI between two images for all possible discrete displacements on a rectangular domain. The output corresponds to a generalization of the, in the field of 1D (EEG) signal analysis, existing notion of cross-mutual information function (CMIF) \cite{pompe1998using} and we therefore use the same name in this work.
We propose a novel approach for efficient computation of CMIF by using cross-correlation (CC) in the frequency domain, providing an output which is equivalent to exhaustively computing MI over all possible discrete displacements.

Our second contribution is a method that combines the fast CMIF computation with a grid search over additional transformation parameters (e.g.\ rotation) to facilitate global multimodal image alignment for transformations with reasonably few degrees of freedom (rigid, affine, etc.).
Figure~\ref{fig:mainillustration} illustrates the main steps of global rigid alignment 
by this method applied on histological images acquired by second harmonic generation (SHG) microscopy, and bright-field (BF) microscopy. 

Our third contribution is a theoretical analysis of the proposed CMIF-algorithm. We show that it is equivalent to a direct histogram-based algorithm for computation of MI. Furthermore, we derive the asymptotic computational complexity of the proposed CMIF-algorithm, and show that it is substantially more computationally efficient than the existing methods.

Our fourth contribution is an empirical evaluation of the run-time of the fast CMIF-algorithm in comparison to the direct histogram-based algorithm. We observe speed-ups ranging from hundreds of times to more than 10,000 times for practically relevant image sizes.

Our fifth contribution is an evaluation of the proposed global rigid alignment method compared with state-of-the-art methods on three distinct and publicly available multimodal rigid alignment benchmark datasets, following the protocol of \cite{lu2021image}. 

The two proposed algorithms are implemented in PyTorch \citep{paszke2019pytorch} for accelerated computation on general-purpose graphics processing units (GPGPUs), and the code is shared as open-source at \codeurl.

\section{MI in image processing}

MI, introduced in \cite{shannon1948mathematical}, is an information-theoretic measure quantifying the mutual dependence between two random variables. It is commonly used in image processing as a similarity measure between images \citep{wells1996multi,maesMultimodalityImageRegistration1997}.

Given two discrete images $A$ and $B$ with domains $\domainof{A} \subset \mathbb{Z}^n$ and $\domainof{B} \subset \mathbb{Z}^n$, intersecting on $\domainof{AB}$, where $\domainof{AB} \neq \emptyset$, with ranges $\rangeof{A}$ and $\rangeof{B}$, MI is defined as
\begin{equation}
I(A, B) = \sum\limits_{a \in \rangeof{A}}{\sum\limits_{b \in \rangeof{B}}{ p^{\,AB}_{\domainof{AB}}(a, b) \log_2{\frac{p^{\,AB}_{\domainof{AB}}(a, b)}{p^{\,A}_{\domainof{AB}}(a)p^{\,B}_{\domainof{AB}}(b)}}}},
\label{eq:mi}
\end{equation}
where $p^{\,AB}_{\domainof{AB}}(a, b)$ denotes the relative frequency of values $a$ and $b$ occurring jointly in $\domainof{AB}$, and $p^{\,A}_{\domainof{AB}}(a)$ denotes the marginal relative frequency of value $a$ occurring in image $A$, within $\domainof{AB}$. The frequencies $p^{\,AB}_{\domainof{AB}}(a, b)$, $p^{\,A}_{\domainof{AB}}(a)$, $p^{\,B}_{\domainof{AB}}(b)$ can be computed using marginal and joint histograms.

Alternatively, MI can be formulated in terms of the marginal and the joint entropies,
\begin{equation}
    I(A, B) = \ha + \hb - \hab\,,
\label{eq:mientropy}
\end{equation}
where the marginal entropy of image $A$, on $\domainof{AB}$ is given by 
\begin{equation}
    \ha = -\sum\limits_{a \in \rangeof{A}}{p^{\,A}_{\domainof{AB}}(a)\log_2{p^{\,A}_{\domainof{AB}}(a)} }\,,
\label{eq:margentropy}
\end{equation}
analogous definition holds for image $B$, and the joint entropy is given by 
\begin{equation}
    \hab = -\sum\limits_{a \in \rangeof{A}}{\sum\limits_{b \in \rangeof{B}}{p^{\,AB}_{\domainof{AB}}(a, b)\log_2{(p^{\,AB}_{\domainof{AB}}(a, b))} }}\,.
\label{eq:jointentropy}
\end{equation}

CMIF describes the MI between two images subject to displacement $\shiftvar \in \mathbb{Z}^n$ (generalization to $n$D of definition in \cite{pompe1998using}),
\begin{equation}
\text{CMIF}(\shiftvar; A, B) = I(A, \shiftimage{B}),
\label{eq:cmif}
\end{equation}
where $\shiftimage{B}$ denotes the image $B$ translated as $\shiftimage{B}(\posvar)=B(\posvar+\shiftvar)$. 
CMIF can also be formulated in terms of entropies, similar to Eq.~\eqref{eq:mientropy}, for entropies defined as functions of displacements, $\ha(\shiftvar)=\ha$, $\hb(\shiftvar)=H_{\shiftimage{B}}$, and $\hab(\shiftvar)=H_{A\shiftimage{B}}$.

\subsection{Image alignment by MI maximization}

An image alignment process based on MI maximization can be expressed as finding a transformation $\widehat{T}$ such that
\begin{equation}
\widehat{T} = \argmax\limits_{T \in \Omega}{I(A, T(B))},
\label{eq:regeq}
\end{equation}
where $A$ is the reference image, and $T(B)$ denotes the geometrically transformed floating image, where transformation $T$ is applied to image $B$ to warp it into the space of $A$, and $\Omega$ is a chosen set of transformations. Solutions to \eqref{eq:regeq} are commonly sought locally, using gradient-based methods \citep{wells1996multi,viola1997alignment,maesMultimodalityImageRegistration1997,knops2006normalized}, or recently through a Gibbs-sampling process \citep{agn2019fast}, or globally through computation of MI for each discrete displacement \citep{barrera2010multimodal}. 
Use of local methods requires a good initial guess since MI typically exhibits a large number of local maxima \cite{pluim2003mutualsurvey}. 
MI can be combined with gradient information to improve the performance when the modalities exhibit similarity of variation \citep{pluim2000image}.
To reduce the dependence on the size of the images and their overlap, a normalized MI (NMI) \citep{studholme1999overlap} was developed. 

Fast computation of MI is critical in many applications, in particular when MI is used as a similarity measure for alignment tasks, due to the large data and number of similarity computations involved. Stochastic sub-sampling is commonly utilized together with gradient-based local optimization methods \cite{viola1997alignment,pluim2003mutualsurvey}. Other works include speeding up local optimization approaches by using GPU-based histogram methods \cite{shams2007speeding,shams2010parallel}.

\subsection{Global optimization over discrete displacements}

Efficient methods for global optimization over discrete displacements (based on fast convolutions performed in the frequency domain or on use of integral images) exist for absolute mean differences, mean square differences \cite{atallah2001faster}, CC and normalized cross-correlation \cite{tsai2003fast}, while (to the best of our knowledge) so far not for MI \citep{paul2021comprehensive, pluim2003mutualsurvey, shams2010parallel, barrera2010multimodal}.

A direct, histogram-based, approach for computing CMIF for all displacements $\shiftvar$ requires, for each $\shiftvar$ (considered in isolation), to iterate over $\domainof{A\shiftimage{B}}$ and increment the joint histogram bin corresponding to the values $a$ and $b$ (and analogously for the marginal histograms).
Finally the entropies (Eq.~\eqref{eq:margentropy} and \eqref{eq:jointentropy}) are computed using relative histograms, which then directly give the MI through Eq.~\eqref{eq:mientropy}. This method is discussed in \cite{barrera2010multimodal}. 
Furthermore,  \cite{shams2007speeding,shams2010parallel} discuss maximization of MI by exhaustive search in the context of their fast approaches for GPU-based histogram computation, and conclude that such an approach is too costly to be useful in practice. 

\section{Method}

We propose a fast and exact algorithm for computing MI for all possible discrete displacements on a rectangular domain (Eq.~\eqref{eq:cmif}), enabling fast global translation-based alignment using MI maximization. We show the equivalence to a direct histogram-based method and we analyze the asymptotic computational complexity. Furthermore, we propose a global multimodal image alignment method based on the proposed CMIF-algorithm.

\subsection{Notation and basic definitions}

Consider pairs of images, ${A\colon \domainof{A} \to \rangeof{A}}$ and ${B\colon \domainof{B} \to \rangeof{B}}$, where ${\domainof{B} \subseteq \domainof{A} \subset \mathbb{Z}^n}$,
together with corresponding given region of interest masks, ${M_A \colon \domainof{A} \to \left\{0, 1\right\}}$ and ${M_B \colon \domainof{B} \to \left\{0, 1\right\}}$ indicating the user-defined part to be included in the computation of MI.
Without loss of generality, we assume that $\domainof{A}$ and $\domainof{B}$ are rectangular subsets of $\mathbb{Z}^n$, since any other subset can be obtained through appropriately defined masks $M_A$ and $M_B$. Let $\domainof{S}$ denote the set of displacements $\shiftvar$ for which the entire domain of the shifted image $B$ intersects with the domain of image $A$. 

The CC, for real-valued functions $f$ and $g$ defined on $\mathbb{Z}^n$, is given by
\begin{equation}
\CC{f}{g} = \sum\limits_{x \in \mathbb{Z}^n}{ f(x)g(x+\shiftvar) }.
\label{eq:cc}
\end{equation}

Let $\binimage{A}{a}$ denote an indicator function (level set) on $\domainof{A}$, of image $A$ being equal to $a \in \rangeof{A}$ and within the mask image $M_A$,
\begin{equation}
\begin{split}
    \binimage{A}{a}(x) =
    \left\{\begin{array}{ll}
        1, & \text{for } \, M_A(x) = 1 \, \text{ and } \, A(x) = a\\
        0, & \text{otherwise}\\    
        \end{array}\right\}.
\end{split}
\end{equation}

\subsection{Algorithm for computing CMIF in the frequency domain}
\label{sec:alg}

Contrary to the direct method, where, given a particular displacement $\shiftvar$, histograms are computed for all image values $a$ and $b$, we instead reorder the required operations; given a single image value $a$, (or) $b$, or a pair of values $(a, b)$, the number of occurrences of the value/pair, and the corresponding entropy contribution, is computed for all $\shiftvar \in \domainof{S}$.

Computation of the joint histogram entries $C^{\,AB}_{a, b}(\shiftvar; A, B)$ can be expressed as
\begin{equation}
C^{\,AB}_{a, b}(\shiftvar; A, B) = \CC{\binimage{A}{a}}{\binimage{B}{b}}\,,
\label{eq:correlationjointhisto}
\end{equation}
and the marginal histogram entries $C^{\,A}_{a}(\shiftvar; A, B)$ and $C^{\,B}_{b}(\shiftvar; A, B)$ respectively, can be expressed as
\begin{equation}
\begin{split}
C^{\,A}_a(\shiftvar; A, B) &= \CC{\binimage{A}{a}}{M_B}\,,\\
C^{\,B}_b(\shiftvar; A, B) &= \CC{M_A}{\binimage{B}{b}}.
\label{eq:correlationmarghisto}
\end{split}
\end{equation}
Complete histograms are computed by evaluating Eq.~\eqref{eq:correlationjointhisto} and Eq.~\eqref{eq:correlationmarghisto} for all image values $a \in \rangeof{A}$ and $b \in \rangeof{B}$.
Another quantity of interest for the CMIF computation is a map $N \colon \domainof{S} \to \mathbb{R}_{\geq 0}$ representing the number of valid points (points where both masks are non-zero) 
for all $\shiftvar \in \domainof{S}$, which enables computation of relative histograms. $N$ is given by
\begin{equation}
\begin{split}
N(\shiftvar; A, B) &= \CC{M_A}{M_B} = \\ \sum\limits_{a\in\rangeof{A}}{C^{\,A}_a(\shiftvar, A, B)} &= \sum\limits_{b\in\rangeof{B}}{C^{\,B}_b(\shiftvar, A, B)}\,.
\label{eq:number}
\end{split}
\end{equation}

The relative histograms obtained after normalizing $\histentrymarga{a}$, $\histentrymargb{b}$, $\histentryjoint{a,b}$ by (pointwise division by) $N$ 
can be inserted as probability functions into the shifted versions of Eq.~\eqref{eq:margentropy} and Eq.~\eqref{eq:jointentropy},
\begin{equation}
\begin{split}
\ha(\shiftvar) &= -\sum\limits_{a \in \rangeof{A}} {\nbinlogn{\frac{C_{a}^{A}(\shiftvar; A, B)}{N(\shiftvar; A, B)}}},\\
\hb(\shiftvar) &= -\sum\limits_{b \in \rangeof{B}} {\nbinlogn{\frac{C_{b}^{B}(\shiftvar; A, B)}{N(\shiftvar; A, B)}}},\\
\hab(\shiftvar) &= -\sum\limits_{a \in \rangeof{A}}\sum\limits_{b \in \rangeof{B}}{ {\nbinlogn{\frac{C_{a,b}^{AB}(\shiftvar; A, B)}{N(\shiftvar; A, B)}}}},
\label{eq:finalentropy}
\end{split}
\end{equation}
which can be used directly to compute CMIF (Eq.~\eqref{eq:cmif}). Rounding $C^{\,A}_{a}, C^{\,B}_{b}, C^{\,AB}_{a,b}$ and $N$ for all $\shiftvar$ to integers we circumvent possible floating-point errors in the computation and reach an exact result. In Eq.~\eqref{eq:finalentropy}, we define that $\frac{0}{0}=0$ and $0 \log_2 0 = 0$.

\subsection{Generalization to spatially weighted MI}

Spatially weighted MI (SWMI) \cite{park2010spatially} is a version of MI that associates each $x \in \domainof{A}$ with a weight, modelling the relative importance of the corresponding image element (and its contribution to the relative frequencies). Our here proposed CMIF-algorithm can be generalized to compute SWMI by replacing the binary mask $M_A$ with a weight mask $W_A\colon \domainof{A} \to \mathbb{R}_{\geq 0}$, and modifying the level-set formulation to a weighted version,
\begin{equation}
L_{a}^{W}\left[A\right](x) =
    \left\{\begin{array}{ll}
        W_A(x), & \text{for } \, A(x) = a\\
        0, & \text{otherwise}\\    
        \end{array}\right\}.
\end{equation}
We may further generalize this approach by considering weight masks for both images $A$ and $B$ where the weight for a given $x \in \domainof{A}$ and $\shiftvar \in \domainof{S}$ is taken to be the product, ${w(x) = W_A(x) \cdot \shiftimage{W_B}\!(x)}$.
        
\subsection{Equivalence of the correlation-based algorithm and the direct method}

The proposed algorithm uses CC to compute CMIF exactly, and identically to the direct method; both involve computing contributions to the same discrete histogram counts $N(\shiftvar, A, B)$, $C^{\,A}_a(\shiftvar, A, B)$, $C^{\,B}_b(\shiftvar, A, B)$ and $C^{\,AB}_{a, b}(\shiftvar, A, B)$ for all $a, b$ and $\shiftvar \in \domainof{S}$. 
The CC (Eq.~\eqref{eq:cc}) computes a sliding inner-product between the two functions. For binary functions (images) $f$ and $g$, Eq.~\eqref{eq:cc} yields the number of elements $x$ where ${f(x)=g(x+\shiftvar)=1}$, for a given $\shiftvar$. 
Hence, the CC of level-set and level-set, level-set and mask, as well as mask and mask, give the histogram entries, for each $\shiftvar$, as given by Eq.~\eqref{eq:correlationjointhisto}, \eqref{eq:correlationmarghisto} and \eqref{eq:number}. 
These inner-products provide all the quantities required for exact computation of the shifted entropies (Eq.~\eqref{eq:finalentropy}), and thus for CMIF. The main distinction between the proposed algorithm and the direct method is the order in which the histogram counts are computed; the direct method computes the histogram entries ($N(\shiftvar; A, B), C^{\,A}_a(\shiftvar, A, B)$, $C^{\,B}_b(\shiftvar, A, B$) for all $a, b$, for a fixed $\shiftvar$, and the proposed approach computes the histogram entries for all $\shiftvar$, for fixed $a, b$, before proceeding.

\subsection{Complexity analysis}

The worst-case run-time complexity of the direct approach is
\begin{equation}
T_D(A, B) = \bigo{ \card{\domainof{B}} \card{\domainof{\searchspace}} + (\card{\rangeof{A}} + \card{\rangeof{B}} + \card{\rangeof{A}} \card{\rangeof{B}}) \card{\domainof{S}}},
\end{equation}
given by its requirement of $\bigo{1}$ work for every element in image $B$ for every $\shiftvar \in \domainof{S}$ and $\bigo{1}$ work per histogram bin. On the other hand, computation of CC in the frequency domain requires, instead, computation of at most $1 + \cardofrangeof{A} + \cardofrangeof{B} + \cardofrangeof{A}\cardofrangeof{B}$ (forward and inverse) DFTs using the Fast Fourier Transform (FFT) algorithm \cite{cooley1965algorithm} and element-wise (complex) multiplications, yielding an asymptotic run-time complexity of
\begin{equation}
T_{F}(A, B) = \mathcal{O}((1+\cardofrangeof{A} + \cardofrangeof{B} + \cardofrangeof{A} \cardofrangeof{B}) \nlogn{\card{\domainof{A}}} )\,.
\label{eq:maincomplexity}
\end{equation}
If $\cardofrangeof{A}$ and $\cardofrangeof{B}$ are treated as (small) constants, the proposed method is asymptotically efficient cf. the direct approach (w.r.t. the size of the images), except for cases with very small \card{\domainof{B}}. 
It has been observed \cite{pluim2000image} that 32 intensity bins provided the best trade-off between (i) flexibility and detail, and (ii) insensitivity to noise, for gradient-based image alignment.

For very small images $B$ or sets of displacements $\card{\domainof{S}}$, or for large value sets $\rangeof{A}$ and $\rangeof{B}$, the direct method may be the better choice. In practice, small or medium sized value sets are usually acceptable and the image sizes are often such that the proposed algorithm is several orders of magnitudes faster. 

\subsection{Method for global multimodal alignment}

We propose a method for image alignment by global MI maximization for transformations with few degrees of freedom (e.g. rigid or affine) by combining the efficient CMIF-algorithm proposed in Sec.~\ref{sec:alg}, with grid search over other transformation parameters such as rotation angle. 
Relying on global search, the method can find the global maximum without smoothing of the images, which otherwise is commonly performed to increase the size of the region of attraction of the global maximum, but may lead to finding a sub-optimal solution \cite{pluim2003mutualsurvey}.

The method, described in Alg.~\ref{alg:reg}, performs global alignment by taking a set of transformations ${\Omega = \left\{T_1, \dots, T_k\right\}}$ (e.g.\ selected from a grid in the parameter space, or through random selection), warping the floating image ($B$) for each transformation using nearest-neighbor (NN) interpolation, computing a CMIF map using the proposed algorithm, and locating the displacement with  highest MI. We use NN interpolation to compute $T(B)$ since $\rangeof{B}$ is assumed to be categorical. A parameter $\gamma\!\in\!\left[0, 1\right]$ controls the required fraction of overlap compared to the maximal observed overlap; $\gamma$ is both used to 
select transformations with sufficient overlap, and to control how large padding is required. Finally, the resulting transformation is taken as the composition of the transformation from $\Omega$ and the $T_{\shiftvar}, \shiftvar\!\in\!X_S$ which leads to the highest MI with required overlap.

The first step of Alg.~\ref{alg:reg} is to quantize the images into suitably small number of levels using some appropriate quantization approach (unless the ranges are already suitable discrete representations). We use $k$-means clustering, which has been shown to yield more efficient utilization of the discrete bins for MI-based image alignment~\cite{knops2006normalized} than equisized binning. Furthermore, $k$-means clustering enables the direct application of the method to multi-channel images, including image pairs with different numbers of channels (denoted $m_A$ and $m_B$). For this work, we use the mini-batch $k$-means algorithm \citep{sculley2010web} which is fast and scalable. We denote $k$ the number of clusters used for quantization of both images, such that $k=\cardofrangeof{A}=\cardofrangeof{B}$.

In Alg.~\ref{alg:reg}, the reference image $A$ and its mask $M_A$ are zero-padded with $\ceil{\card{\domainof{B}}_i (1-\gamma)}$ units (before and after) along (each) axis $i$, where $\card{\domainof{B}}_i$ denotes the size (diameter) of the (rectangular) domain of image $B$ along axis $i$.

\begin{algorithm}[t]
\caption{
Image alignment by global MI maximization}\label{alg:reg}
\begin{algorithmic}[1]
\small
 \renewcommand{\algorithmicrequire}{\textbf{Input:}}
 \renewcommand{\algorithmicensure}{\textbf{Output:}}
 \Require $A \colon \domainof{A} \to \mathbb{R}^{m_A}, M_A\colon\domainof{A}\to\{0,1\},
 \newline
 B \colon \domainof{B} \to \mathbb{R}^{m_B}, M_B\colon\domainof{B}\to\{0,1\},
\;\; \Omega \in \left\{T_1, \dots, T_k\right\},\; \gamma \in \left[0, 1\right]$.
 \Ensure $I \in \mathbb{R}_{\geq 0}, \widehat{T} \colon \mathbb{R}^n \to \mathbb{R}^n$.\smallskip
    \State $A \gets \Call{QUANTIZE}{A}, B \gets \Call{QUANTIZE}{B}$
    \State $A, M_A \gets \Call{ZERO-PAD}{A, M_A; \domainof{B}, 1-\gamma}$
    \State $I \gets 0, \widehat{T} \gets {\textbf {IDENTITY}}$
    \For{$T \in \Omega$}
        \State $\mathcal{N} \gets \gamma \max\limits_{\shiftvar \in \domainof{S}} N(\shiftvar; A, T(B))$
        \State $\shiftvar \gets \argmax\limits_{\hspace*{-5mm}\shiftvar \in \domainof{S} \text{ s.t. } N(\shiftvar; A, T(B)) \geq \mathcal{N}\hspace*{-8mm}}{\text{CMIF}(\shiftvar, A, T(B))}$ 
        \State $I_T \gets\text{CMIF}(\shiftvar, A, T(B))$
        \If{$I_T > I$}
            \State $I \gets I_T, \widehat{T} \gets T \circ T_{\shiftvar}$
        \EndIf
    \EndFor
\end{algorithmic}
\end{algorithm}

For the task of global rigid alignment we run the method twice: (i) first using grid search to reach a coarse alignment at an angle $\theta$, (ii) followed by an (optional) refinement step, where random search is employed in an interval around the best angle found by the first grid search, sampling uniformly from the interval $\left[\theta - \frac{2\pi}{\card{\Omega}}, \theta + \frac{2\pi}{\card{\Omega}}\right]$, where $\card{\Omega}$ denotes the number of angles in the first grid search stage.

Both image alignment tasks, where the reference and floating images are similar in size, as well as patch retrieval (template matching) tasks, where the reference image is larger than the floating image, can be efficiently solved directly by Alg.~\ref{alg:reg}.

\section{Implementation}

We implement both the proposed CMIF-algorithm, as well as the global image alignment procedure (Alg.~\ref{alg:reg}) in Python/PyTorch as a way to utilize parallel processing on a GPU. Using Python $3.8.8$ and PyTorch $1.8.1$, the method runs almost entirely on the GPU, including image warping, padding, level-sets, and CC ($\text{FFTs}$ and complex multiplication).

Mini-batch $k$-means clustering relies on the provided implementation in the {\texttt{sklearn}} package \citep{scikit-learn}, version 0.24.1.

\section{Performance analysis}

We consider three datasets (\cite{luDatasetsEvaluationMultimodal2021}): (i) aerial images \cite{volpiSemanticSegmentationUrban2015} with infrared (IR) as one modality and color images as the other, (ii) cytological images \cite{vicarQuantitativePhaseDynamicsApoptosis2020} with quantitative phase images as one modality and fluorescence images as the other, (iii) histological images \cite{keikhosraviIntensitybasedRegistrationBrightfield2020} with SHG images as one modality and BF as the other, (illustrated in Fig.~\ref{fig:mainillustration}). The aerial image dataset and cytological image dataset consist of $864$ and $5040$ images respectively, and are both divided into three distinct folds; the histological image dataset consists of a single set of 536 images. The aerial and cytological images are of size $300 \times 300$, and the histological images are of size $834 \times 834$.

\subsection{Run-time analysis}

We investigate the run-times of the proposed CMIF-algorithm, compared to a direct histogram-based algorithm, implemented in Python/PyTorch, using the built-in PyTorch implementation for computing image histograms on the GPU. The algorithms implemented in PyTorch are very similar to methods 1 and 2 in \citep{shams2010parallel}, where the method is chosen dynamically based on the number of requested bins, and thus enabling a fair comparison of the proposed method and the direct method. The sort-and-count algorithm for computing the histograms  \citep{shams2010parallel} has been shown to be relevant mostly for larger number of bins, and therefore we do not consider it here, given that we observe that small number of bins is shown to work well in this global optimization context, as seen in Sec.~\ref{sec:rigidresults}.

\subsubsection{Experimental setup}

We select one of the histological image pairs at random, and crop/pad it to various sizes such that the side-lengths of the reference image are a power of two $\left\{128, 256, 512, 1024, 2048, 4096\right\}$ and the other (floating) image is of half the size along each dimension, which is the scenario encountered when aligning equally sized images with Alg.~\ref{alg:reg} and $\gamma=0.5$. 
We compute the CMIF-map for all $\shiftvar \in X_S$. 

Experiments are run on a Nvidia GeForce GTX 2080 GPU, with 11 GB of memory.

We also apply the image alignment method (Alg.~\ref{alg:reg}) on cytological and histological images, and measure the run-time for a number of configurations, including $50$, $100$, and $200$ grid steps (with 32 additional refinement angles), with $k \in \left\{8, 16, 32\right\}$, with $k$-means batch size of $1000$ and max iterations $25$.

\subsubsection{Results}

\begin{figure}[t]
\centering
\includegraphics[width=1.0\columnwidth,height=3.2cm]{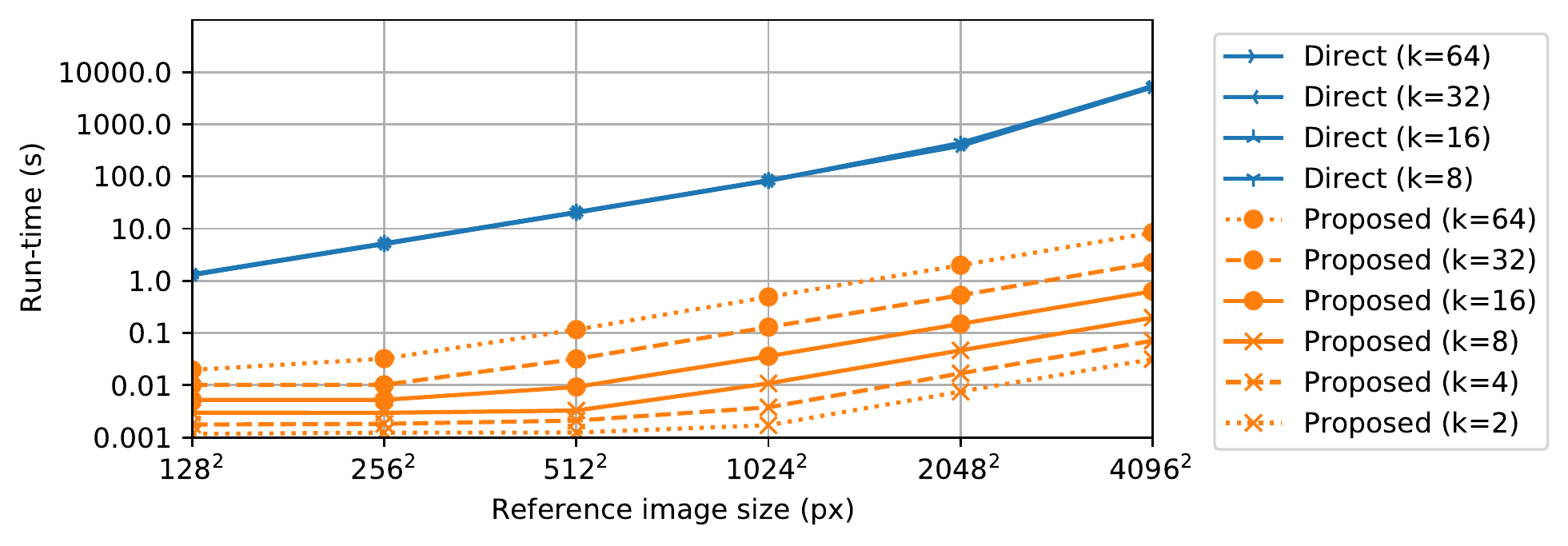}
\caption{Run-time of the proposed CMIF-algorithm for various reference image sizes and choices of $k$, in comparison to the direct histogram-based method, implemented in PyTorch and executed on a GPU. We observe that the proposed CMIF-algorithm is several orders of magnitude faster than the direct method. 
The empirical run-times of the proposed algorithm exhibits a strong dependency on $k$, as expected from Eq.~\eqref{eq:maincomplexity}.
The empirical run-times for the direct method are relatively stable w.r.t. $k$; we observed at most a difference of $30\%$ between such choices for $k \in \left\{8, 16, 32, 64\right\}$.
}
\label{fig:timeresults}
\end{figure}

The results of the run-time performance experiment regarding computation of a complete CMIF-map are summarized in Fig.~\ref{fig:timeresults}. We observe that the proposed algorithm is substantially (between 100 times to more than 10000 times) faster than the direct method, and the difference is particularly noteworthy for low numbers of bins, $k \in \left\{2, 4, 8\right\}$.

Table~\ref{tab:regtimes} shows the run-time for the global rigid alignment method for a number of realistic configurations. For the smaller cytological images, with 8 bins in particular, we observe performance compatible with real-time applications. One rigid image alignment for the histological dataset using the direct method with $k=16$ and 200 angles + 32 refinement angles, runs in approximately 19 hours (compared to $32.9$s with the fast CMIF-algorithm).

\begin{table}[tb]
\centering
\caption{Run-time in seconds for global rigid image alignment for a number of configurations using images taken from the cytological and histological dataset.}
\label{tab:regtimes}
    \resizebox{1.0\columnwidth}{!}{
    \begin{tabular}{cc|ccc}
    \multicolumn{2}{l|}{\textbf{Cytological}} & \multicolumn{3}{c}{\textbf{$k$}} \\
    & & 8 & 16 & 32 \\ \hline
    \parbox[t]{2mm}{\multirow{3}{*}{\rotatebox[origin=c]{90}{\textbf{Angles}}}} & \Tstrut 50 & 0.8 & 2.2 & 7.3 \\
    & 100 & 1.2 & 3.4 & 11.4 \\
    & 200 & 2.2 & 6.1 & 22.1 \\
    \end{tabular} \qquad
    \begin{tabular}{cc|ccc}
    \multicolumn{2}{l|}{\textbf{Histological}} & \multicolumn{3}{c}{\textbf{$k$}} \\
    & & 8 & 16 & 32 \\ \hline
    \parbox[t]{2mm}{\multirow{3}{*}{\rotatebox[origin=c]{90}{\textbf{Angles}}}} & \Tstrut 50 & 4.0 & 11.8 & 42.2 \\
    & 100 & 6.1  & 18.9 & 67.9  \\
    & 200 & 10.3  & 32.9 & 119 \\
    \end{tabular}
}
\end{table}

\subsection{Rigid image alignment}

To evaluate the efficacy of the proposed image alignment method, compared to several existing methods, we follow the evaluation protocol of \cite{lu2021image}, comprising a benchmark for evaluating the effectiveness of 2D rigid image alignment methods on three distinct datasets. 
The study includes comparison of four image-to-image (I2I) translation methods, as well as a state-of-the-art contrastive learning method for modality transfer, which are all used to attempt to transform the multimodal alignment task into an easier monomodal alignment task, and are then combined with either a high-performance intensity-based method \citep{ofverstedtFastRobustSymmetric2019} or a well-known feature-based method \citep{loweObjectRecognitionLocal1999}. Finally, local optimization (gradient-based) MI is also included; it is applied to multimodal data directly.
Even though 2D rigid transformation is among the simplest models, solving these alignment tasks still poses a challenge in the presence of multiple distinct modalities. It is also a realistic task in microscopy and aerial settings where relative scale between the images may be known a priori, or readily estimated, while arbitrary rotations can be encountered.

\subsubsection{Experimental setup}

Following \citep{lu2021image}, we apply the proposed alignment method on all three datasets. The performance measure is based on the average Euclidean distance between image corners considered as landmarks in the reference image space, and the corresponding recovered landmarks of the aligned floating image. An alignment is considered successful if the error is less than $2\%$ of the width of the images \citep{lu2021image}.

An important parameter of the proposed method is $k$, which has a large impact on the run-time (Eq.~\ref{eq:maincomplexity}) while also directly influencing how detailed structures can be represented in the quantized representations. We run the alignment task for all the images in the included datasets (aerial, cytological, and histological), with $k \in \left\{2, 4, 8, 16, 32, 64\right\}$, with $k$-means batch size of $100$ and max iterations $100$, and measure the success-rates. 

The number of angles for the grid search, as well as if including the refinement step or not, are also important considerations. We evaluate the method on the three datasets using a set of angle counts $\left\{25, 50, 100, 150, 200, 250, 300\right\}$, with and without refinement with $32$ randomly selected angles,
and measure the success-rate for each configuration.

We use circular masks, to avoid bias from presence or absence of the signal contents in the corners of the images.

\subsubsection{Results}
\label{sec:rigidresults}

\begin{figure}[t]
\centering
\includegraphics[width=1.0\columnwidth]{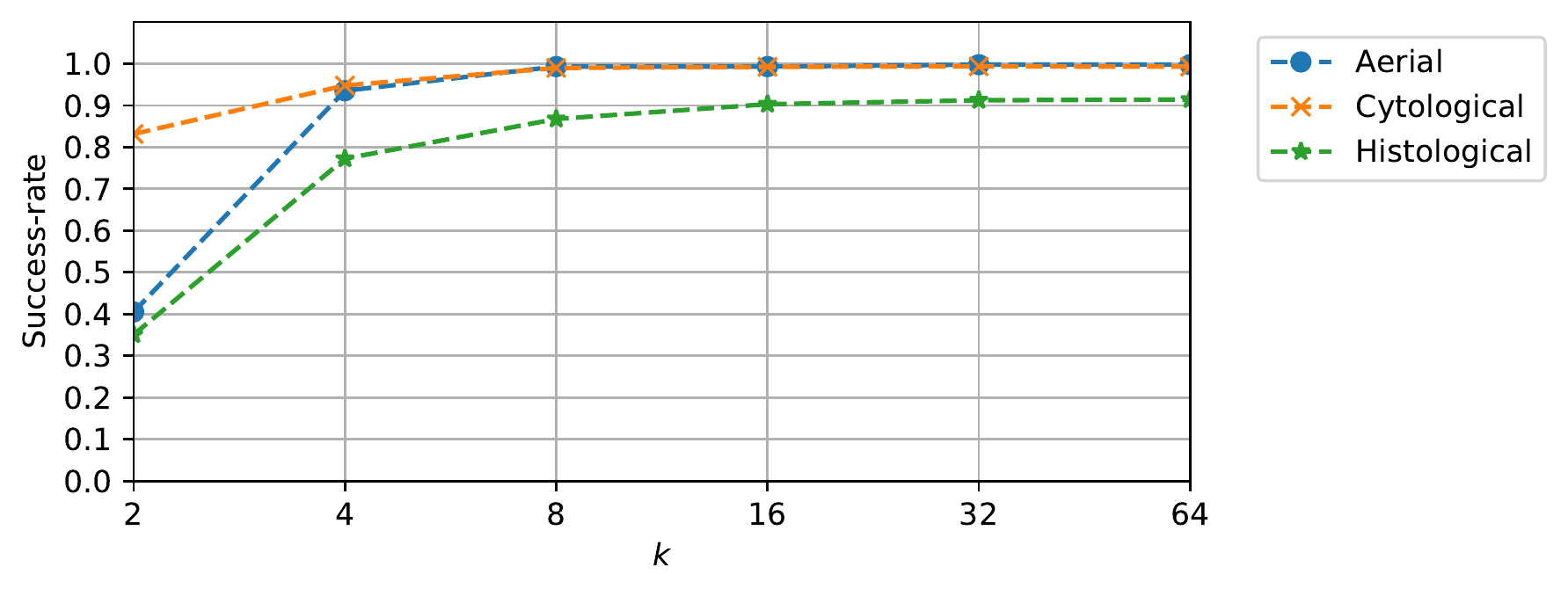}
\caption{The relationship between success-rate and choice of $k$ on the three evaluation datasets. }
\label{fig:moreresults}
\end{figure}

\begin{figure}[t]
\centering
\includegraphics[width=1.0\columnwidth]{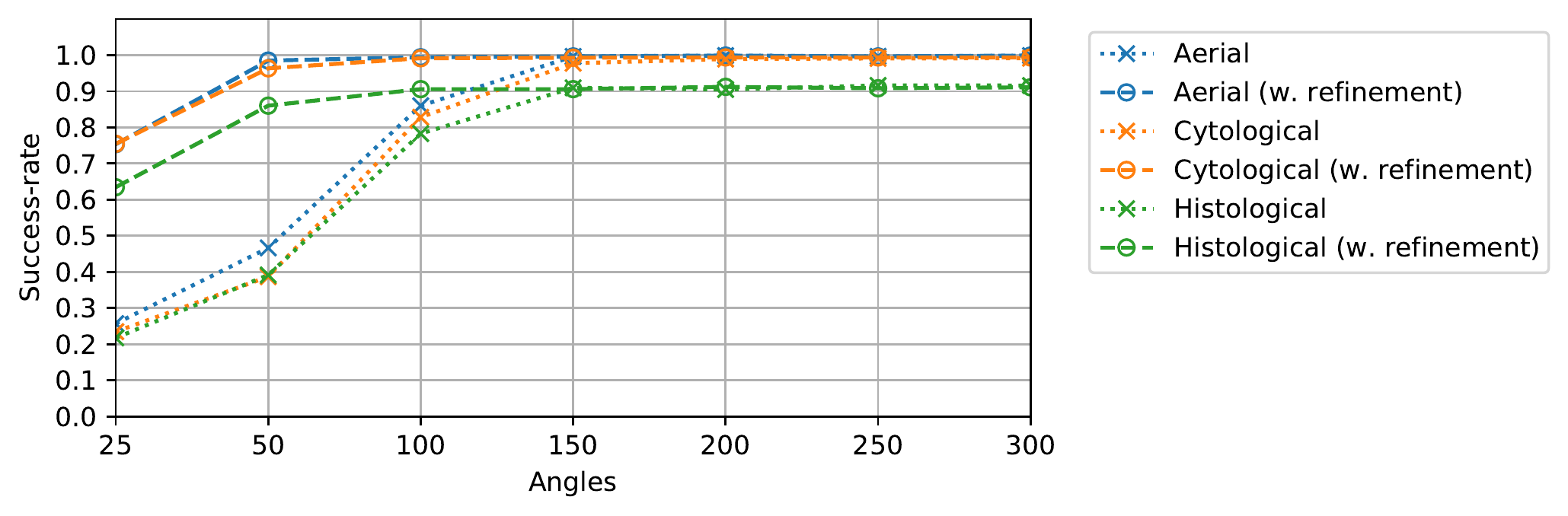}
\caption{The relationship between success-rate and the number of angles in the grid search, using $k=32$, both with, or without, the subsequent refinement step. In both cases, the performance increases substantially up to 150 angles for the three datasets.}
\label{fig:angleplot}
\end{figure}

First we present the performance on the three evaluation datasets (as measured by success-rates) for different choices of: $k$ in Fig.~\ref{fig:moreresults}, and angle count in Fig.~\ref{fig:angleplot}; The success-rate increases substantially up to $k=16$, and $150$ angles.

Furthermore, we present the performance of the considered methods in comparison to a number of alternative methods in Tab.~\ref{tab:regresults}. The four I2I translation-based methods are combined into a single row, where the maximum success-rate along each column is shown. The proposed method is the best choice on two of the datasets (the cytological and the histological) where it exhibits a substantial improvement over existing methods. CoMIR combined with SIFT reaches 100\% success-rate on the Zurich dataset, while the proposed method delivers a success-rate of 99.8\%, which is better than all included intensity-based methods.

\begin{table}[tb]
    \centering
    \caption{Success rates (presented as percentages) of the evaluated methods on rigid alignment tasks on three datasets. Results on the aerial and cytological datasets are presented as empirical mean $\pm$ std-dev over the three folds. Bold marks the best result on each dataset. Used parameters for the proposed method are $k=32$, $\gamma=0.5$, 200 equispaced angles in $\left[-\pi, \pi\right]$ plus 32 random refinement angles. For details about the reference methods, and their configuration, see \citep{lu2021image}.}
    \label{tab:regresults}
    \resizebox{\columnwidth}{!}{%
    \begin{tabular}{cc@{\;\;}cc@{\;\;}cc@{\;}c}
        \hline
        \textbf{Dataset} & \multicolumn{2}{c}{\textbf{Aerial Data}} & \multicolumn{2}{c}{\textbf{Cytological Data}} & \multicolumn{2}{c}{\textbf{Histological Data}} \\ \hline
        \textbf{Method} & $\alpha$-AMD & SIFT          & $\alpha$-AMD  & SIFT         & $\alpha$-AMD   & SIFT   \\ \hline
        I2I & $80.2\pm3.9$ & $98.3\pm0.5$ & $71.1\pm5.8$ & $24.4\pm6.2$ & $28.4$ & $0$ \\ \hline
        CoMIR & 91.8$\pm$7.7 & {\bf 100.0$\pm$0.0} & 68.0$\pm$14.0 & 72.5$\pm$7.1 &  81.3 & 59.3 \\ \hline
        Local MI               & \multicolumn{2}{c}{68.9$\pm$3.8}         & \multicolumn{2}{c}{89.9$\pm$3.0}              & \multicolumn{2}{c}{47.8}                       \\ \hline
        \textbf{Proposed} & \multicolumn{2}{c}{$99.8\pm0.4$} & \multicolumn{2}{c}{{\bf 99.4$\pm$0.4}} & \multicolumn{2}{c}{{\bf 91.2}} \\ \hline
    \end{tabular}%
    }
\end{table}

\section{Discussion and Conclusion}

We present a novel fast CMIF-algorithm which computes MI for all discrete displacements of two images (similarly as cross-correlation). The algorithm works in $n$D. The proposed algorithm enables fast global alignment (for transformation models with few degrees of freedom), 
has very few parameters, and is straightforward to configure for new applications.
Quantization of image ranges with $k$-means clustering provides efficient use of a limited number of quantization levels, and enables aligning images with different numbers of channels.

We compare the run-time of the proposed CMIF-algorithm with a direct histogram-method (using a fast GPU-based histogram algorithm), and observe speed-ups ranging from hundreds of times to more than 10,000 times for practically relevant image sizes, indicating the potential for real-time global multimodal alignment. The presented methods are highly parallelizable; the work can readily be distributed over multiple GPUs or computers. 

Furthermore, we evaluate the performance of the proposed alignment method on three datasets, and compare it with both gradient-based MI method, and recent methods which rely on deep learning. We observe excellent performance of the proposed method on all three datasets, and we conclude that it is the overall top performing method, even outperforming state of the art deep learning-based methods. Finally, the proposed method does not require aligned image pairs (or any training), and have few parameters to tune, which are advantages in comparison to the deep learning-based methods.

One limitation of the proposed method is that it does not support cost-effective sub-pixel alignment. Future work could involve application of a suitable refinement step based on local optimization towards this end.

\section*{Declaration of competing interests}

The authors declare  no competing interests. 

\section*{Acknowledgments}

This work was supported by the Wallenberg AI, Autonomous Systems and Software Program (WASP) AI-Math initiative; VINNOVA (MedTech4Health project 2017-02447); and the Swedish Research Council (project 2017-04385).

\bibliographystyle{model1-num-names}
\bibliography{refs}

\clearpage

\end{document}